\documentclass[11pt]{article}

\usepackage[letterpaper,margin=1in]{geometry}
\usepackage{times}

\usepackage{amsmath,amsfonts,bm}

\def\eqref#1{equation~\ref{#1}}

\def\1{\bm{1}}

\DeclareMathAlphabet{\mathsfit}{\encodingdefault}{\sfdefault}{m}{sl}
\SetMathAlphabet{\mathsfit}{bold}{\encodingdefault}{\sfdefault}{bx}{n}

\newcommand{\E}{\mathbb{E}}

\usepackage[numbers]{natbib}

\usepackage{hyperref}
\usepackage{url}

\usepackage{amssymb,amsthm,mathtools}
\usepackage{microtype}
\usepackage{booktabs,multirow,array,tabularx,adjustbox}
\usepackage{graphicx}
\usepackage{xcolor}
\usepackage[nameinlink,capitalise,noabbrev]{cleveref}
\usepackage{placeins}
\allowdisplaybreaks

\newcommand{\cA}{\mathcal A}
\newcommand{\cD}{\mathcal D}

\newcommand{\cI}{\mathcal I}

\newcommand{\norm}[1]{\left\lVert #1\right\rVert}

\newcommand{\tableformat}{\centering\renewcommand{\arraystretch}{1.05}\setlength{\tabcolsep}{4pt}}

\newtheorem{theorem}{Theorem}[section]
\newtheorem{proposition}[theorem]{Proposition}
\newtheorem{corollary}[theorem]{Corollary}

\theoremstyle{definition}

\theoremstyle{remark}

\title{When Does Forecasting Reveal Temporal Structure? A Stability Analysis of Time-Series Structural Selection}

\author{
Qipeng Qian$^{1,2}$ \qquad Yuntao Qian$^{2}$\\[4pt]
$^{1}$SUPCON Technology, Hangzhou, China\\
$^{2}$College of Artificial Intelligence, Zhejiang University, Hangzhou 310027, China\\[4pt]
\texttt{qianqipeng@supcon.com} \qquad
\texttt{ytqian@zju.edu.cn}
}

\date{}

\begin{document}
\maketitle

\begin{abstract}
Forecast accuracy is often used as a proxy for temporal structure discovery,
but predictive performance and structural identifiability are not equivalent.
Different temporal mechanisms can achieve similar forecast errors, while small
forecast differences may still contain sufficient information for recovery.
In this work, we study when forecast-only structural selection can be trusted.
We show that a vanishing forecast margin does not necessarily imply structural
ambiguity, and establish a stability perspective that evaluates structural
separation relative to uncertainty in the selection objective. This perspective
provides both a sufficient condition for reliable selection and a continuous
measure of selection difficulty.
Experiments across controlled and end-to-end settings demonstrate that forecast
margin alone is insufficient, while the proposed stability measure better
characterizes when forecast-based structural selection succeeds or fails. Our results suggest that predictive accuracy should be treated as evidence for
structure discovery only when its separation is sufficiently robust.
\end{abstract}

\section{Introduction}

A low forecast MSE answers one question: how well does the model predict? It does not by itself answer a second question: did the forecasting procedure select the temporal structure that generated the data? The distinction matters whenever the reported delay or temporal relation is itself used for interpretation, diagnosis, or decision making.

Time series make these two questions easy to confuse. Nearby lags can be highly correlated, and past observations may provide alternative predictive information. Therefore, forecast accuracy alone may not reliably indicate whether the underlying temporal structure has been correctly identified. However, it remains unclear how forecast separation relates to structural recoverability and when forecast-based selection can be trusted.

We separate the problem into three questions:
\[
\boxed{
\text{recoverability}
\;\longrightarrow\;
\text{forecast margin}
\;\longrightarrow\;
\text{structural stability}
}.
\]
\textbf{Recoverability} asks whether the full observed trajectory contains enough information to distinguish the true temporal structure from the candidate alternatives. The \textbf{forecast margin} is the increase in the best achievable per-observation forecast MSE when an alternative structure is used. The \textbf{structural stability} question is different again: after training and finite-sample evaluation, does the objective used to choose among candidate structures still rank the true structure first?

Forecast similarity and structural identifiability are fundamentally different notions. Even in a simple point-delay model, the forecast gap between two candidate structures can become arbitrarily small while the full trajectory still contains enough information to recover the true structure.
Thus, predictive indistinguishability does not necessarily imply structural ambiguity.

This observation leads to a stability perspective for forecast-based
structural selection. The forecast margin alone is not sufficient to characterize whether a selected structure is reliable; what matters is whether the remaining structural separation is large relative to the uncertainty of the empirical selection objective. We formalize this relationship through the stability ratio, which provides both a sufficient condition for correct selection and a continuous measure of selection difficulty. Controlled and end-to-end experiments verify that this ratio better characterizes structural selection reliability than the forecast margin alone.

Our contributions are summarized as follows:
\begin{itemize}
\item We identify and characterize the gap between forecast performance and temporal structural recoverability. We show that a vanishing forecast margin does not imply structural ambiguity, because the full trajectory can still contain sufficient information for recovery. 

\item We establish when forecast-based structural selection can be trusted. The stability ratio compares structural separation with uncertainty in the selection objective, providing both a sufficient condition for reliable selection and a quantitative measure of selection difficulty.

\item We validate this perspective through controlled and end-to-end experiments, showing when forecast-only structural selection succeeds or fails and demonstrating the limitations of forecast margin alone. 
\end{itemize}

\section{Related Work}

Related work relevant to this paper can be organized around three questions:
whether temporal structures are recoverable, how predictors identify useful temporal information, and whether predictive performance provides reliable
evidence about underlying mechanisms.

\paragraph{Delay recovery and temporal structure.}
Delay and lead--lag estimation are classical problems in system identification and time-series analysis \citep{bjorklund2003review,sadler2006survey,ljung1999system,ma2023datadriven,cattaldo2024variable,heyse2021lag,zhao2024lift,runge2019earthcausation,runge2019pcmci,runge2020pcmcip}. More recent work studies whether temporal relations are recoverable from the observed input support. In particular, \citet{kuskova2026recoverability} give a pre-fit diagnostic for interaction recoverability under dependent lagged inputs. Our recoverability question is close to this line of work, but our main focus is the relation between recoverability, forecast margin, and the objective used for structural selection.

\paragraph{Learning temporal dependence for forecasting.}
Time-series models use masks, selected context, attention, and bottlenecks to learn which parts of the history matter for prediction \citep{ozyegen2022interpretability,liu2024contralsp,liu2024timexpp,zhang2025mew}. IB-Forecast learns explicit input masks \citep{zheng2026ibforecast}, while DSPR learns regime-dependent history together with interaction structure \citep{zhang2026dspr}. Forecast-necessity testing removes candidate temporal relations and checks how much forecast accuracy changes \citep{kuskova2026necessity}. These methods ask which history a predictor uses or needs. We instead start from a known candidate structure and measure how strongly the forecast objective separates it from alternatives.

\paragraph{Prediction accuracy and model structure.}
Different models can have similar forecast MSE while using different variables or decision rules. This issue appears in work on underspecification, predictive multiplicity, shortcut learning, and explanation faithfulness \citep{damour2022underspecification,marx2020predictive,geirhos2020shortcut,jain2019attention,jacovi2020faithful,ross2017right}. Time-series dependence makes this especially important because nearby lags and past outputs can replace part of the same predictive signal \citep{tunyi2026marginal}. Existing approaches study temporal dependence
recovery or predictive importance separately, but they do not characterize when a forecast objective provides reliable evidence for the underlying temporal structure. We address this gap by formulating temporal structural selection
through three related quantities: recoverability from the full trajectory,
forecast separation between candidate structures, and structural stability of the selection objective.

Together, these lines of work motivate our study of when forecast-based
selection can be trusted for temporal structure discovery.

\section{Problem Setup}\label{sec:setup}

We use two controlled mechanisms:
\begin{align}
\text{P1:}\quad
&Y_t=\alpha U_{t-d}+\epsilon_t,\quad \epsilon_t\sim\mathcal N(0,\sigma^2);
\label{eq:p1-model}\\
\text{P2:}\quad
&Y_t=aY_{t-1}+bU_{t-d}+\epsilon_t.
\label{eq:p2-model}
\end{align}
P1 isolates a single input delay. P2 adds output memory and lets an alternative delay re-fit the other coefficients. This makes P2 the main setting for predictive substitution.

Let $L$ be the largest candidate lag and define
\[
\cI_L=\{L+1,\dots,T\},
\qquad
n_{\rm eff}=T-L.
\]
For a candidate delay $d$,
\[
S_du=(u_{t-d})_{t\in\cI_L}
\]
is the shifted input trajectory.

For P2 we use the shared window
\begin{equation}
\cI_L^{\rm P2}
=
\{L+2,\ldots,T\},
\qquad
n_{\rm P2}=T-L-1,
\label{eq:p2-corrected-window}
\end{equation}
so that every candidate uses the same observed history. The exact assumptions are listed in Appendix~\ref{app:assumptions}.

\section{Theory: Recoverability, Forecast Margin, and Stability}\label{sec:theory}

The theory answers three questions in the order in which they arise. First, if two delays are possible, how well can the full trajectory tell them apart? Second, how different are their best forecast MSEs? Third, even if the true structure has the lower ideal risk, when is that ordering preserved by the objective actually used for selection?

\subsection{P1: prediction can be accurate while structure remains identifiable}

We begin with the simplest possible setting. The purpose of P1 is not to study a difficult forecasting model. It is to separate two questions that are often treated as the same question:

\begin{enumerate}
    \item \textbf{Recovery:} given the whole observed trajectory, can we determine which delay generated the data?
    \item \textbf{Forecasting:} if we use a candidate delay for prediction, how much larger is its prediction error?
\end{enumerate}

These two questions use different information. Recovery uses all observations jointly to make one decision about the hidden delay. Forecasting evaluates the average prediction error at each time step. The proposition below shows that these two quantities can behave differently.

Suppose the true delay is either $d_0$ or $d_1$, with equal prior probability. For a fixed input trajectory $u$, both candidates generate Gaussian output trajectories with the same covariance. Their difference is only the mean trajectory:
\[
\mu_{d_0}=\alpha S_{d_0}u,
\qquad
\mu_{d_1}=\alpha S_{d_1}u .
\]
Therefore the amount of information available for recovering the delay is
\begin{equation}
D_{\rm KL}(d_0\Vert d_1\mid u)
=
\frac{\alpha^2}{2\sigma^2}
\|S_{d_0}u-S_{d_1}u\|_2^2 .
\label{eq:p1-kl-main}
\end{equation}

The optimal recovery procedure uses the whole trajectory and chooses the delay with the larger likelihood. Its error probability is
\begin{equation}
e_{\rm rec}^\star(d_0,d_1;u)
=
\Phi\!\left(
-\sqrt{D_{\rm KL}(d_0\Vert d_1\mid u)/2}
\right),
\label{eq:p1-bayes-main}
\end{equation}
where $\Phi$ is the standard normal cumulative distribution function.

Now consider the forecasting view. The true delay gives the conditional mean used to generate the data, so its per-observation forecast MSE is only the irreducible noise: 
\[
r_{n_{\rm eff}}(d_0)=\sigma^2 .
\]
The alternative delay makes a different prediction at every step. We define its excess forecast loss as
\begin{equation}
\Delta_{n_{\rm eff}}(d_1;d_0,u)
:=
r_{n_{\rm eff}}(d_1)-r_{n_{\rm eff}}(d_0)
=
\frac{\alpha^2}{n}
\|S_{d_0}u-S_{d_1}u\|_2^2 .
\label{eq:p1-forecast-margin}
\end{equation}

Importantly, $\Delta_{n_{\rm eff}}$ is not the forecast MSE. It is only the \emph{forecast advantage of the true structure over the alternative structure}. The following proposition connects the recovery problem and the forecasting problem.

\begin{proposition}[P1: identifiable structure with vanishing forecast advantage]
\label{prop:p1-tradeoff}
Under P1 with independent Gaussian noise of variance $\sigma^2$ and equal prior probability on $d_0,d_1$,
\begin{align}
D_{\rm KL}(d_0\Vert d_1\mid u)
&=
\frac{n_{\rm eff}\Delta_{n_{\rm eff}}(d_1;d_0,u)}{2\sigma^2},
\label{eq:p1-info-margin}\\
e_{\rm rec}^\star(d_0,d_1;u)
&=
\Phi\!\left(
-\frac{\sqrt{n_{\rm eff}\Delta_{n_{\rm eff}}(d_1;d_0,u)}}{2\sigma}
\right),
\label{eq:structural-error-eta}\\
r_{n_{\rm eff}}(d_0)&=\sigma^2,\nonumber\\
r_{n_{\rm eff}}(d_1)&=\sigma^2+\Delta_{n_{\rm eff}}(d_1;d_0,u).
\label{eq:tradeoff-identity}
\end{align}

Consider a sequence of problems indexed by $n$ such that
\[
\Delta_{n_{\rm eff}}\rightarrow0,
\qquad
n_{\rm eff}\Delta_{n_{\rm eff}}\rightarrow\infty,
\]
then we have 
\[
e_{\rm rec}^\star\rightarrow0,
\qquad
r_{n_{\rm eff}}(d_1)-r_{n_{\rm eff}}(d_0)\rightarrow0.
\]
The proof is shown in Appendix \ref{app:proof prop p1-tradeoff}. 
\end{proposition}

The proposition describes a separation between two notions of being “good”.

The first one is \textbf{structural identifiability}. Although the difference between the two delays at each individual time step becomes smaller, the same small difference is observed repeatedly over $n$ observations. The accumulated evidence therefore increases:
\[
D_{\rm KL}(d_0\Vert d_1\mid u)
=
\frac{n_{\rm eff}\Delta_{n_{\rm eff}}}{2\sigma^2}\rightarrow\infty ,
\]
and the optimal delay recovery error goes to zero.

The second one is \textbf{forecast preference}. The wrong delay is still a wrong structure, but its additional prediction loss is only $\Delta_{n_{\rm eff}}$. When $\Delta_{n_{\rm eff}}\rightarrow0$, the two structures become almost indistinguishable if they are compared only through average forecast MSE. 

Therefore, the proposition does not say that the wrong delay becomes correct. It says that a wrong temporal structure can become nearly as good for forecasting while the full trajectory still contains enough information to recover the true structure. This distinction motivates the rest of the paper: forecast performance alone does not determine whether a learned temporal structure is correct.

\subsection{P2: the forecast margin after the wrong delay is re-fitted}

The P1 analysis compares two delay candidates while keeping the model form
fixed. In practice, however, a candidate delay is always evaluated after the
remaining model parameters are re-fitted. Therefore, the relevant forecast comparison is not the original margin in P1, but the margin that remains after the alternative delay has been optimized. 

Consider the data-generating process of P2 in \cref{eq:p2-model}, for a candidate delay $d$, let
\[
r^\star(d)
:=\inf_{a,b}\mathcal r(d,a,b)
\]
denote the best forecast MSE achievable by this delay after re-fitting the regression coefficients. We define the 
\emph{profiled forecast margin} 
\[
\Delta_{\rm prof,n_{\rm P2}}(d)
:=r^\star(d)-r^\star(d^\star).
\]
This is the P2 counterpart of the P1 forecast margin: it measures the remaining
forecast disadvantage of the wrong delay after the alternative structure has
been given its best possible coefficients.

The profiled forecast margin also determines the separation between the full trajectory distributions. Let $P^\star$ be the true trajectory distribution and let $P_{d,a,b,\sigma^2}$ be the Gaussian trajectory distribution generated by candidate delay $d$ with coefficients $(a,b)$ and the true innovation variance.
Since $(a,b)$ are unknown under the candidate structure, we minimize the KL
separation over them:
\[
J_{\rm prof}^{\rm fixed}(d^\star,d)
:=
\inf_{a,b}
D_{\rm KL}(P^\star\Vert P_{d,a,b,\sigma^2}).
\]
If the candidate is also allowed to estimate its own innovation variance, we
similarly define
\[
J_{\rm prof}^{\rm free}(d^\star,d)
:=
\inf_{a,b,s^2}
D_{\rm KL}(P^\star\Vert P_{d,a,b,s^2}).
\]

\begin{proposition}[P2: re-fitting changes the forecast margin and preserves its trajectory accumulation]
\label{prop:arx-profile-margin}

For any alternative delay $d$, the profiled forecast margin and the remaining
trajectory separation satisfy
\begin{align}
J_{\rm prof}^{\rm fixed}(d^\star,d)
&=
\frac{n_{\rm P2}\Delta_{\rm prof,n_{\rm P2}}(d)}{2\sigma^2},
\label{eq:p2-fixed-profile-margin}
\\
J_{\rm prof}^{\rm free}(d^\star,d)
&=
\frac{n_{\rm P2}}{2}
\log\!\left(
1+
\frac{\Delta_{\rm prof,n_{\rm P2}}(d)}{\sigma^2}
\right).
\label{eq:p2-free-profile-margin}
\end{align}
Therefore, re-fitting changes only the size of the forecast margin: the original
margin in P1 is replaced by the smaller remaining margin after optimization.
The same margin is still accumulated over the full trajectory through the factor
$n_{\rm P2}$. The proof is shown in Appendix \ref{app:proof prop arx-profile-margin}. 
\end{proposition}

The proposition gives the P2 analogue of the P1 result. The alternative delay is
stronger than in P1 because it is allowed to adjust its regression coefficients.
As a result, part of the predictive difference can be absorbed by re-fitting,
and the remaining forecast margin can become much smaller. However, the
trajectory-level separation is still determined by the accumulated quantity
$n_{\rm P2}\Delta_{\rm prof,n_{\rm P2}}(d)$ rather than by the per-step margin alone.
Thus, the same phenomenon as in P1 can occur after re-fitting: the wrong delay
can become almost as good for forecasting while the full trajectory still
contains increasing evidence that the delay is different.

The mechanism that reduces the profiled margin is analyzed in Appendix~\ref{app:partialcorr}. In particular, predictive substitution through past outputs or alternative lags can reduce the remaining forecast difference after re-fitting.

\subsection{From the ideal forecast margin to the objective used for selection}

So far, every candidate has been compared by its ideal candidate risk. 
In practice, the delay is not selected using the ideal risk $r^\star(d)$.
Instead, each candidate is fitted from finite data and evaluated using an
empirical forecast loss. The resulting objective $\widehat r(d)$ can deviate
from $r^\star(d)$ and may select a different delay. 
Thus, the questions discussed in the part are: when the observed objective changes the ranking of candidates and how large $|r^\star(d)-\widehat r(d)|$ can be before the selection ranking becomes incorrect.

Define the \emph{total objective distortion} 
\begin{equation}
\tau_{\rm total}
:=
\max_{d}
\left|
\widehat r(d)-r^\star(d)
\right|.
\label{eq:total-objective-distortion}
\end{equation}
Let the minimum ideal forecast margin be
\begin{equation}
\Delta^{\min}
:=
\min_{d\ne d^\star}
\left[
r^\star(d)-r^\star(d^\star)
\right]
>0.
\label{eq:generic-margin}
\end{equation}
For P2, this is the minimum profiled forecast margin.

The following theorem gives a direct condition under which the ordering cannot change.

\begin{theorem}[A margin larger than twice the total objective distortion preserves the true structure]
\label{thm:structural-stability}
If
\begin{equation}
\Delta^{\min}>2\tau_{\rm total},
\label{eq:stability-condition}
\end{equation}
then $d^\star$ is the unique minimizer of $\widehat r$. The factor $2$ is sharp in the worst case: if $\Delta^{\min}<2\tau_{\rm total}$, an allowed distortion can make an alternative structure have a smaller objective than the true one; at equality it can create a tie. The proof is shown in Appendix \ref{app:thm structural-stability}. 
\end{theorem}
The theorem gives a simple condition for when forecast performance can guarantee structural correctness. Suppose a method is only known to be within $\varepsilon$ of the best ideal candidate risk
\[
r^\star(\hat d)
\leq
r^\star(d^\star)+\varepsilon.
\]
Then the method identifies the true structure only when 
\[
\varepsilon<\Delta^{\min}.
\]
Thus, near-optimal forecast MSE is therefore not, by itself, evidence of structural correctness. 

We then define \emph{stability ratio}
\begin{equation}
S_{\rm total}
:=
\frac{\Delta^{\min}}{2\tau_{\rm total}}.
\label{eq:stability-ratio}
\end{equation}
$S_{\rm total}>1$ guarantees correct structural selection. When $S_{\rm total}\le1$, the selected structure can be either correct or wrong.

\section{Experiments}
\label{sec:experiments}

\subsection{Forecast-only structural selection protocol}
\label{sec:exp setup}

This section describes the common forecast-only structural-selection procedure
used in all experiments. For each candidate structure, a predictor is fitted
using only the information allowed by that structure, and the candidate is
selected by its empirical forecast MSE.

For a candidate delay $d$, let $\widehat f_d$ denote the fitted predictor.
An independent selection set is used to compute empirical forecast MSE 
\[
\widehat r(d)
=
\frac1n
\sum_{i=1}^{n}
\left(
Y_i-\widehat f_d(X_i)
\right)^2 ,
\]
and the selected structure is
\begin{equation}
\widehat d
=
\arg\min_d \widehat r(d).
\label{eq:forecast-only-selection}
\end{equation}

The true delay is never used during training, model selection, or structural
selection. It is only used after selection to evaluate recovery accuracy.

We consider two predictor classes. The neural predictor is a small MLP whose
inputs are restricted to the history permitted by each candidate delay. The
linear baseline is ordinary least squares (OLS) using the same candidate
inputs. Since the P2 data-generating process is linear, OLS provides a
well-specified baseline and separates structural-selection effects from
additional nonlinear modeling capacity.

The following experiments examine different consequences of the theoretical
analysis: recovery versus forecast separation in P1, predictive substitution
in P2, stability under finite-sample estimation, and the behavior of learned
predictors. 

A semi-synthetic industrial-process illustration with controlled temporal
structures is provided in Appendix~\ref{app:semi-synthetic-industrial}. This additional experiment uses realistic process dynamics while keeping the temporal structure known by construction.

\subsection{P1: recovery remains possible as the forecast-MSE gap vanishes}
\label{sec:p1-experiment}

This experiment demonstrates one of the central phenomena studied in this
paper: a structural variable can remain recoverable even when the per-observation
forecast-MSE difference becomes asymptotically negligible.

We consider the local sequence
\[
\alpha_n=c\,n^{-\beta/2},
\]
with $\beta=0.5$, which satisfies the regime
\[
\Delta_n\rightarrow 0,
\qquad
n\Delta_n\rightarrow\infty .
\]
The first condition implies that the forecast risk difference between competing
structures vanishes at the individual observation level. The second condition
ensures that the accumulated trajectory-level information still diverges,
allowing consistent structural recovery.

\begin{figure}[htbp]
    \centering

    \includegraphics[width=0.48\linewidth]{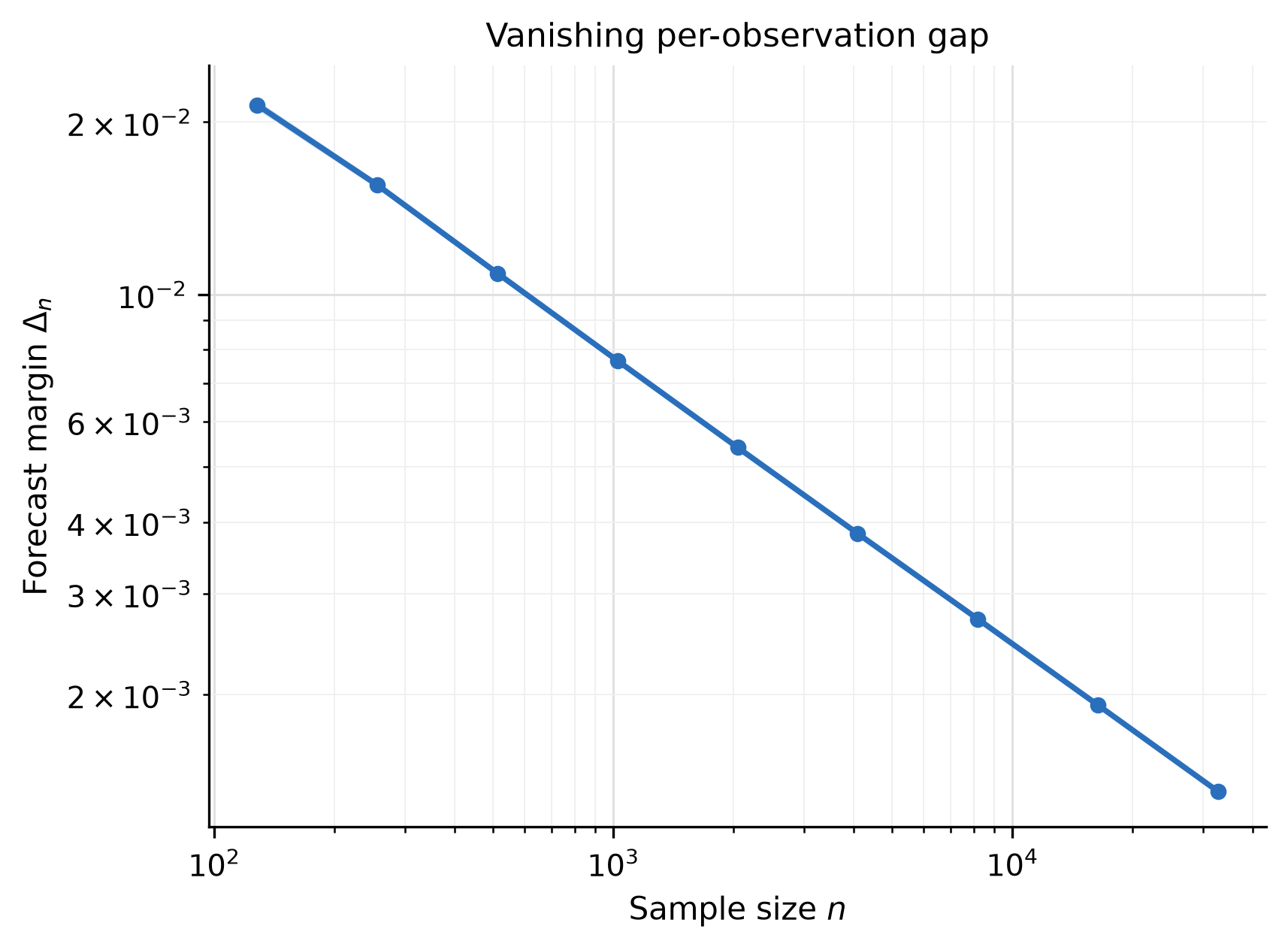}
    \hfill
    \includegraphics[width=0.48\linewidth]{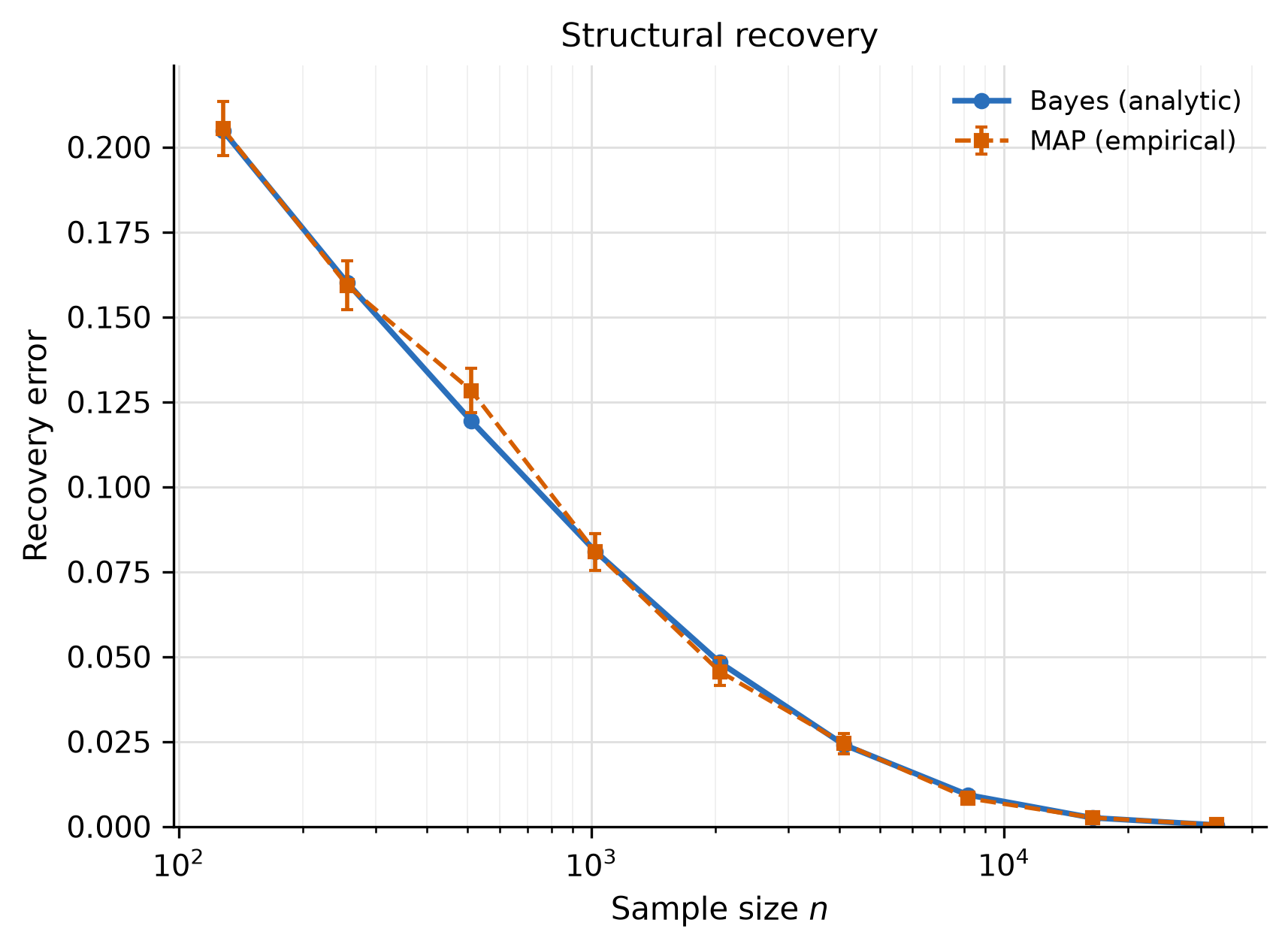}

    \caption{
    P1 experiment: forecast separation vanishes while structural recovery
    remains possible. 
    (a) The forecast margin $\Delta_n$ decreases with sample size. 
    (b) Structural recovery error decreases toward zero. 
    }
    \label{fig:p1-recovery}
\end{figure}

Figure~\ref{fig:p1-recovery} illustrates this behavior. In panel (a), the
forecast margin $\Delta_n$ decreases continuously as the sample size increases,
showing that the two candidate structures become increasingly difficult to
distinguish using a single forecast error. Nevertheless, panel (b) shows that
the corresponding recovery error decreases toward zero. The analytic Bayes
error and the empirical MAP recovery error closely match throughout the entire
range, confirming that the observed recovery behavior agrees with the
information-theoretic prediction.

Therefore, P1 provides a concrete example where diminishing predictive
separation does not imply loss of structural identifiability. Instead, the
relevant quantity for recovery is the accumulated information over the full
trajectory rather than the forecast-MSE gap of an individual observation.

\subsection{The $2\tau_{\rm total}$ boundary is a stability guarantee}
\label{sec:stability-experiment}

The next experiment isolates Theorem~\ref{thm:structural-stability} from model
training. We generate $450{,}000$ independent P1 candidate-objective samples.
Because the ideal candidate risks are known exactly, we can compute
$\tau_{\rm total}$ directly for every sample and check whether the selected
structure changes.

\begin{table}[htbp]
\tableformat
\caption{Finite-objective stability study. $S_{\rm total}>1$ is the sufficient
region from Theorem~\ref{thm:structural-stability}.}
\label{tab:objective-stability}
\begin{tabular}{lc}
\toprule
Quantity & Result \\
\midrule
Independent objective samples & 450,000 \\
Samples with $S_{\rm total}>1$ & 97,238 \\
Wrong selections with $S_{\rm total}>1$ & 0 \\
Wrong-selection rate with $S_{\rm total}\le1$ & 19.48\% \\
Spearman correlation across fixed $S_{\rm total}$ bins & -0.857 \\
\bottomrule
\end{tabular}
\end{table}

No sample above the boundary ($S_{\rm total}>1$) selects the wrong structure.
Below the boundary ($S_{\rm total}\le1$), some selections are wrong and many
are still correct. This is exactly the one-way statement of
Theorem~\ref{thm:structural-stability}: the sufficient condition guarantees
stability, but violating the condition does not imply failure.

\begin{figure}[htbp]
    \centering
    \includegraphics[width=0.6\linewidth]{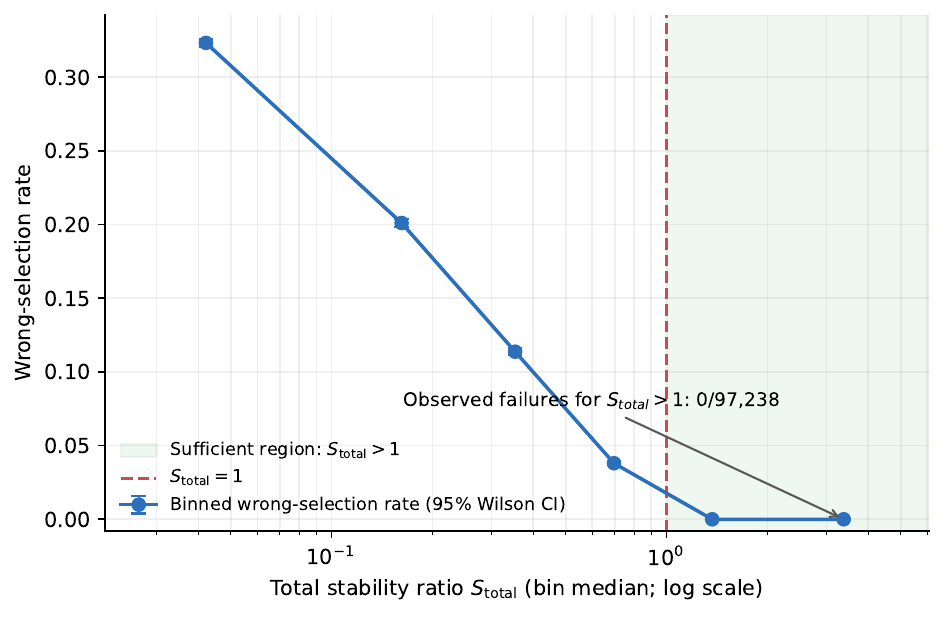}
    \caption{
    Continuous relationship between the stability ratio and structural
    selection difficulty. Samples are grouped by $S_{\rm total}$ and the
    wrong-selection rate is reported with 95\% Wilson confidence intervals.
    The dashed line indicates the theoretical boundary
    $S_{\rm total}=1$. No wrong selections are observed in the sufficient
    region $S_{\rm total}>1$.
    }
    \label{fig:stability-ratio-curve}
\end{figure}

Beyond this binary guarantee, the stability ratio also provides a continuous
measure of selection difficulty. Figure~\ref{fig:stability-ratio-curve}
shows the wrong-selection rate after grouping samples by their stability
ratio. The selection error decreases as $S_{\rm total}$ increases, while the
sufficient region $S_{\rm total}>1$ contains no observed failures. This illustrates that the stability ratio can serve not only as a boundary condition for guaranteed selection, but also as a quantitative measure of selection difficulty under finite-objective uncertainty.

\subsection{End-to-end stress test: structural selection follows the stability ratio}
\label{sec:stress-test}

The previous experiments isolate different components of forecast-only
structural selection. P1 shows that a small forecast margin does not
necessarily prevent structural recovery. P2 explains why practical
forecast-only selection operates on a profiled margin after predictive
substitution and parameter re-optimization. The stability theorem then
identifies the relevant quantity as the separation margin relative to the
total objective distortion.

This experiment evaluates whether the same principle holds in a complete
forecast-only learning pipeline. We construct a broader P2 stress regime where
the structural-selection problem is not artificially easy. The grid spans
\[
\Delta_{\rm prof}\in[0.00589,0.25],
\]
and varies training size, selection size, input correlation, output memory,
noise level, and input gain. It contains $3{,}456$ frozen neural candidate
models and matched OLS candidate fits. Each trained model is evaluated under
three independent selection-set sizes, resulting in $5{,}184$ neural and
$5{,}184$ OLS structural selections.

The goal is not to compare neural and linear predictors, but to test whether
structural failures are governed by the proposed stability ratio. The similar
overall error rates of the two predictors indicate that the observed behavior
is a property of the selection problem rather than a specific model class.

\begin{table}[htbp]
\tableformat
\caption{End-to-end stress test of the stability ratio. The table evaluates
both the theoretical boundary and the continuous relationship between the
stability ratio and structural-selection difficulty.}
\label{tab:p2-stress}
\begin{tabular}{lcc}
\toprule
Quantity & Neural & OLS \\
\midrule
\multicolumn{3}{c}{\textbf{Boundary verification}}\\
\midrule
Structural selections & 5,184 & 5,184 \\
Wrong-selection rate & 38.46\% & 37.50\% \\
Cases with $S_{\rm total}>1$ & 84 & 96 \\
Wrong selections with $S_{\rm total}>1$ & 0 & 0 \\
Median $\tau_{\rm total}$ & 0.8109 & 0.7906 \\
\midrule
\multicolumn{3}{c}{\textbf{Continuous difficulty measure}}\\
\midrule
Spearman correlation:
wrong rate vs. $S_{\rm total}$ & -0.905 & -- \\
Point-biserial correlation:
wrong selection vs. $\log\Delta_{\rm prof}$ & -0.0817 & -- \\
Point-biserial correlation:
wrong selection vs. $\log S_{\rm total}$ & -0.1975 & -- \\
\bottomrule
\end{tabular}
\end{table}

The sufficient condition from Theorem~\ref{thm:structural-stability} is
recovered in the full learning setting: no wrong structural selections occur
when $S_{\rm total}>1$. 
Below this boundary, failures become possible but are not guaranteed, consistent with the one-way nature of the theoretical guarantee. 

The stability ratio also provides a continuous measure of selection difficulty
beyond the binary guarantee. As summarized in Table~\ref{tab:p2-stress}, the wrong-selection rate decreases monotonically with increasing $S_{\rm total}$, with a Spearman correlation of $-0.905$. Moreover, $S_{\rm total}$ is more informative than the profiled forecast margin alone ($-0.0817\&-0.1975$)
This confirms that structural difficulty is governed not by the forecast
margin alone, but by the margin relative to total objective distortion.

Additional controlled analyses on why the forecast margin alone is insufficient are provided in Appendix~\ref{sec:margin-analysis}.

\section{Discussion and Conclusion}

Forecasting and structural identification are often treated as aligned
objectives, but our results show that they answer fundamentally different
questions. Forecasting evaluates whether a candidate structure can achieve
accurate predictions, whereas structural selection asks whether the underlying
temporal mechanism can be distinguished from competing alternatives. Our main
finding is that forecast accuracy alone is not sufficient evidence for
structural correctness. Reliable forecast-only structural selection requires
separating recoverability, effective forecast separation, and objective
stability.

A key implication of our analysis is that forecast separation alone does not
characterize structural identifiability. A small forecast margin does not
necessarily imply structural ambiguity: the forecast difference can vanish
while the accumulated information over a trajectory remains sufficient for
structural identification. Moreover, the separation observed by a forecasting
model may differ from the underlying structural difference because candidate
structures can retain predictive similarity through alternative temporal
patterns and model adaptation.

The central conclusion from these results is that structural selection depends
on the forecast margin relative to the uncertainty of the empirical selection
objective. The condition $\Delta^{\min}>2\tau_{\rm total}$ formalizes this requirement: a candidate structure can be reliably selected only when the forecast separation is larger than the variation introduced by finite-sample estimation and model optimization. The experiments confirm this principle
across controlled and end-to-end settings: the sufficient stability boundary is never violated, and the stability ratio provides a more informative measure of selection difficulty than the forecast margin alone.

For applications in which the temporal structure itself is important, forecast accuracy should therefore not be used as a stand-in for structural evidence. A reliable evaluation should distinguish recoverability, forecast separation, and uncertainty in the selection objective. These quantities answer different questions, and the stability ratio connects predictive separation with selection reliability when a structure is chosen by forecast performance.

\clearpage

\subsection*{AI use statement}
Large language models were used as general-purpose research assistance for language polishing, identifying potentially relevant related work, and generating portions of the experimental code. References surfaced with AI assistance were checked by the authors against the original sources before citation. AI-generated code used in the reported experiments was manually reviewed, tested, and checked against the declared experimental protocol before execution. Reported numerical results were obtained by executing the reviewed code rather than by asking a language model to generate result values. The authors take full responsibility for the paper's claims, citations, code, and experimental results.

\subsection*{Reproducibility statement}
The appendix states the assumptions, exact risk and information formulas, proof details, data-generation rules, and experimental protocols. The reported experiments use fixed grids and fixed random seeds, with separate checks for population calculations, sampling, and structural selection. Structural selection uses forecast MSE only. The release records resolved configurations, seed maps, model hashes, candidate-wise selection losses, analytic margins, total objective distortion, and stability ratios.

\clearpage
\bibliography{iclr2027_conference}

@book{ljung1999system,
  author    = {Ljung, Lennart},
  title     = {System Identification: Theory for the User},
  edition   = {2},
  publisher = {Prentice Hall PTR},
  address   = {Upper Saddle River, NJ},
  year      = {1999},
  isbn      = {9780136566953}
}

@INPROCEEDINGS{bjorklund2003review,
  author={Bjorklund, S. and Ljung, L.},
  booktitle={42nd IEEE International Conference on Decision and Control (IEEE Cat. No.03CH37475)}, 
  title={A review of time-delay estimation techniques}, 
  year={2003},
  volume={3},
  number={},
  pages={2502-2507 Vol.3},
  doi={10.1109/CDC.2003.1272997}}

@inproceedings{sadler2006survey,
  author    = {Sadler, Brian M. and Kozick, Richard J.},
  title     = {A Survey of Time Delay Estimation Performance Bounds},
  booktitle = {2006 IEEE Sensor Array and Multichannel Signal Processing Workshop Proceedings},
  pages     = {282--288},
  year      = {2006},
  address   = {Waltham, MA, USA},
  publisher = {IEEE},
  doi       = {10.1109/SAM.2006.1706138}
}

@article{heyse2021lag,
  author  = {Heyse, Jolan and Sheybani, Laurent and Vulliémoz, Serge and van Mierlo, Pieter},
  title   = {Evaluation of Directed Causality Measures and Lag Estimations in Multivariate Time-Series},
  journal = {Frontiers in Systems Neuroscience},
  volume  = {15},
  pages   = {620338},
  year    = {2021},
  doi     = {10.3389/fnsys.2021.620338}
}

@article{runge2019earthcausation,
  author  = {Runge, Jakob and Bathiany, Sebastian and Bollt, Erik and Camps-Valls, Gustau and Coumou, Dim and Deyle, Ethan and Glymour, Clark and Kretschmer, Marlene and Mahecha, Miguel D. and Muñoz-Marí, Jordi and van Nes, Egbert H. and Peters, Jonas and Quax, Rick and Reichstein, Markus and Scheffer, Marten and Schölkopf, Bernhard and Spirtes, Peter and Sugihara, George and Sun, Jie and Zhang, Kun and Zscheischler, Jakob},
  title   = {Inferring Causation from Time Series in Earth System Sciences},
  journal = {Nature Communications},
  volume  = {10},
  pages   = {2553},
  year    = {2019},
  doi     = {10.1038/s41467-019-10105-3}
}

@article{runge2019pcmci,
  author  = {Runge, Jakob and Nowack, Peer and Kretschmer, Marlene and Flaxman, Seth and Sejdinovic, Dino},
  title   = {Detecting and Quantifying Causal Associations in Large Nonlinear Time Series Datasets},
  journal = {Science Advances},
  volume  = {5},
  number  = {11},
  pages   = {eaau4996},
  year    = {2019},
  doi     = {10.1126/sciadv.aau4996}
}

@inproceedings{runge2020pcmcip,
  author    = {Runge, Jakob},
  title     = {Discovering Contemporaneous and Lagged Causal Relations in Autocorrelated Nonlinear Time Series Datasets},
  booktitle = {Proceedings of the 36th Conference on Uncertainty in Artificial Intelligence},
  series    = {Proceedings of Machine Learning Research},
  volume    = {124},
  pages     = {1388--1397},
  year      = {2020},
  editor    = {Peters, Jonas and Sontag, David},
  publisher = {PMLR},
  url       = {https://proceedings.mlr.press/v124/runge20a.html}
}

@article{ma2023datadriven,
  title={Data-driven approach for time-delay estimation of industrial processes},
  author={Ma, Xin-Yue and Huang, Chun-Qing},
  journal={ISA Transactions},
  volume={137},
  pages={35--58},
  year={2023},
  doi={10.1016/j.isatra.2023.01.028}
}

@article{cattaldo2024variable,
  author  = {Cattaldo, Marco and Ferrer, Alberto and M{\aa}ge, Ingrid},
  title   = {Variable time delay estimation in continuous industrial processes},
  journal = {Chemometrics and Intelligent Laboratory Systems},
  volume  = {246},
  pages   = {105082},
  year    = {2024},
  doi     = {10.1016/j.chemolab.2024.105082},
  url     = {https://doi.org/10.1016/j.chemolab.2024.105082}
}

@inproceedings{jain2019attention,
  title     = {Attention is not Explanation},
  author    = {Jain, Sarthak and Wallace, Byron C.},
  booktitle = {Proceedings of NAACL-HLT},
  pages     = {3543--3556},
  year      = {2019},
  doi       = {10.18653/v1/N19-1357}
}

@inproceedings{jacovi2020faithful,
  title     = {Towards Faithfully Interpretable {NLP} Systems: How Should We Define and Evaluate Faithfulness?},
  author    = {Jacovi, Alon and Goldberg, Yoav},
  booktitle = {Proceedings of the 58th Annual Meeting of the Association for Computational Linguistics},
  pages     = {4198--4205},
  year      = {2020},
  doi       = {10.18653/v1/2020.acl-main.386}
}

@misc{ross2017right,
      title={Right for the Right Reasons: Training Differentiable Models by Constraining their Explanations}, 
      author={Andrew Slavin Ross and Michael C. Hughes and Finale Doshi-Velez},
      year={2017},
      eprint={1703.03717},
      archivePrefix={arXiv},
      primaryClass={cs.LG},
      url={https://arxiv.org/abs/1703.03717}, 
}

@inproceedings{zhao2024lift,
  author    = {Zhao, Lifan and Shen, Yanyan},
  title     = {Rethinking Channel Dependence for Multivariate Time Series Forecasting: Learning from Leading Indicators},
  booktitle = {The Twelfth International Conference on Learning Representations},
  year      = {2024},
  url       = {https://proceedings.iclr.cc/paper_files/paper/2024/hash/b52b07a239a7afa155ca25cf17a55074-Abstract-Conference.html}
}

@misc{kuskova2026recoverability,
      title={When Are Neural Interaction Discoveries Real? Identifiability, Recoverability, and a Pre-Fit Diagnostic}, 
      author={Valentina Kuskova and Dmitry Zaytsev and Michael Coppedge},
      year={2026},
      eprint={2606.08390},
      archivePrefix={arXiv},
      primaryClass={cs.LG},
      url={https://arxiv.org/abs/2606.08390}, 
}

@article{ozyegen2022interpretability,
  author  = {Ozyegen, Ozan and Ilic, Igor and Cevik, Mucahit},
  title   = {Evaluation of Interpretability Methods for Multivariate Time Series Forecasting},
  journal = {Applied Intelligence},
  volume  = {52},
  number  = {5},
  pages   = {4727--4743},
  year    = {2022},
  doi     = {10.1007/s10489-021-02662-2}
}

@inproceedings{zhang2025mew,
  author    = {Zhang, Jiahui and Zhou, Zhengyang and Du, Wenjie and Wang, Yang},
  title     = {Enhancing the Maximum Effective Window for Long-Term Time Series Forecasting},
  booktitle = {Advances in Neural Information Processing Systems},
  volume    = {38},
  year      = {2025},
  url       = {https://openreview.net/forum?id=Gmwsy7TlFI}
}

@article{zheng2026ibforecast,
  author  = {Zheng, Xu and Cheng, Wei and Chen, Zhuomin and Sha, Mo and Ni, Jingchao and Luo, Dongsheng},
  title   = {Information Bottleneck Learning for Faithful Time Series Forecasting Explanations},
  journal = {arXiv preprint arXiv:2607.28124},
  year    = {2026},
  doi     = {10.48550/arXiv.2607.28124},
  url     = {https://arxiv.org/abs/2607.28124}
}

@article{tunyi2026marginal,
  author  = {Tunyi, Amadeo},
  title   = {The Failures of Marginal Influence-Based Attribution Methods for Global Time Series Explanations},
  journal = {arXiv preprint arXiv:2607.16236},
  year    = {2026},
  doi     = {10.48550/arXiv.2607.16236},
  url     = {https://arxiv.org/abs/2607.16236}
}

@inproceedings{liu2024contralsp,
  author    = {Liu, Zichuan and Zhang, Yingying and Wang, Tianchun and Wang, Zefan and Luo, Dongsheng and Du, Mengnan and Wu, Min and Wang, Yi and Chen, Chunlin and Fan, Lunting and Wen, Qingsong},
  title     = {Explaining Time Series via Contrastive and Locally Sparse Perturbations},
  booktitle = {The Twelfth International Conference on Learning Representations},
  year      = {2024},
  url       = {https://openreview.net/forum?id=qDdSRaOiyb}
}

@inproceedings{liu2024timexpp,
  author    = {Liu, Zichuan and Wang, Tianchun and Shi, Jimeng and Zheng, Xu and Chen, Zhuomin and Song, Lei and Dong, Wenqian and Obeysekera, Jayantha and Shirani, Farhad and Luo, Dongsheng},
  title     = {{TimeX++}: Learning Time-Series Explanations with Information Bottleneck},
  booktitle = {Proceedings of the 41st International Conference on Machine Learning},
  pages     = {32062--32082},
  year      = {2024},
  volume    = {235},
  series    = {Proceedings of Machine Learning Research},
  publisher = {PMLR},
  url       = {https://proceedings.mlr.press/v235/liu24bl.html}
}

@article{kuskova2026necessity,
   title={Beyond coefficients: Forecast-necessity testing for interpretable causal discovery in nonlinear time-series models},
   author={Kuskova, Valentina and Zaytsev, Dmitry and Coppedge, Michael},
   ISSN={2334-0754},
   url={http://dx.doi.org/10.32473/flairs.39.1},
   DOI={10.32473/flairs.39.1},
   number={1},
   volume    = {39},
   journal={The International FLAIRS Conference Proceedings},
   publisher={University of Florida George A Smathers Libraries},
   year={2026},
   month=May }

@article{damour2022underspecification,
  author  = {D'Amour, Alexander and Heller, Katherine and Moldovan, Dan and Adlam, Ben and Alipanahi, Babak and Beutel, Alex and Chen, Christina and Deaton, Jonathan and Eisenstein, Jacob and Hoffman, Matthew D. and Hormozdiari, Farhad and Houlsby, Neil and Hou, Shaobo and Jerfel, Ghassen and Karthikesalingam, Alan and Lucic, Mario and Ma, Yian and McLean, Cory and Mincu, Diana and Mitani, Akinori and Montanari, Andrea and Nado, Zachary and Natarajan, Vivek and Nielson, Christopher and Osborne, Thomas F. and Raman, Rajiv and Ramasamy, Kim and Sayres, Rory and Schrouff, Jessica and Seneviratne, Martin and Sequeira, Shannon and Suresh, Harini and Veitch, Victor and Vladymyrov, Max and Wang, Xuezhi and Webster, Kellie and Yadlowsky, Steve and Yun, Taedong and Zhai, Xiaohua and Sculley, D.},
  title   = {Underspecification Presents Challenges for Credibility in Modern Machine Learning},
  journal = {Journal of Machine Learning Research},
  volume  = {23},
  number  = {226},
  pages   = {1--61},
  year    = {2022},
  url     = {https://jmlr.org/papers/v23/20-1335.html}
}

@inproceedings{marx2020predictive,
  author    = {Marx, Charles and Calmon, Flavio and Ustun, Berk},
  title     = {Predictive Multiplicity in Classification},
  booktitle = {Proceedings of the 37th International Conference on Machine Learning},
  series    = {Proceedings of Machine Learning Research},
  volume    = {119},
  pages     = {6765--6774},
  year      = {2020},
  publisher = {PMLR},
  url       = {https://proceedings.mlr.press/v119/marx20a.html}
}

@article{geirhos2020shortcut,
  author  = {Geirhos, Robert and Jacobsen, J{\"o}rn-Henrik and Michaelis, Claudio and Zemel, Richard and Brendel, Wieland and Bethge, Matthias and Wichmann, Felix A.},
  title   = {Shortcut Learning in Deep Neural Networks},
  journal = {Nature Machine Intelligence},
  volume  = {2},
  pages   = {665--673},
  year    = {2020},
  doi     = {10.1038/s42256-020-00257-z}
}

@misc{zhang2026dspr,
      title={DSPR: Dual-Stream Physics-Residual Networks for Trustworthy Industrial Time Series Forecasting}, 
      author={Yeran Zhang and Pengwei Yang and Guoqing Wang and Tianyu Li},
      year={2026},
      eprint={2604.07393},
      archivePrefix={arXiv},
      primaryClass={cs.LG},
      url={https://arxiv.org/abs/2604.07393}, 
}
\bibliographystyle{iclr2027_conference}

\clearpage
\appendix

\section{Detailed Assumptions and Notation}
\label{app:assumptions}

\paragraph{Candidate and input contract.}
The candidate delay is defined by the data-generating process. In the controlled P1 experiments, the input law does not depend on the candidate delay. Exact Bayes statements condition on every observed boundary variable used in the likelihood.

\paragraph{Noise contract.}
P1 uses independent Gaussian noise. P2 assumes $\epsilon_t$ is independent of the past output and observed exogenous input. The exact P2 statements use the same shared history under every candidate.

\paragraph{P2 boundary window.}
P2 uses \cref{eq:p2-corrected-window}. This removes the first transition whose required past output is not included in the shared observed history.

\paragraph{Profiled P2 risk.}
For each candidate delay, the other linear coefficients are re-fitted before the candidate risks are compared. The resulting difference is the profiled forecast margin. Profile-KL is used as a population separation quantity; it is not called an exact composite-hypothesis Bayes recovery error.

\paragraph{Local-sequence contract.}
Statements with $\Delta_{n_{\rm eff}}\to0$ and $n_{\rm eff}\Delta_{n_{\rm eff}}\to\infty$ describe a sequence of statistical problems indexed by $n$. Parameters such as $\alpha_n$ or $b_n$ may change with $n$. For one fixed identifiable stationary system, the per-observation forecast margin need not go to zero as more observations are collected.

\paragraph{Forecast-only selection contract.}
Training, checkpoint choice, and structural selection do not use the true delay label. Candidate structure is selected by forecast MSE on an independent selection set. The ideal profiled candidate risks are used only after selection to evaluate $\tau_{\rm total}$ and $S_{\rm total}$.

\section{Proofs}
\label{app:proofs}

The proof dependency follows the logical hierarchy of the paper:
\begin{enumerate}
    \item information identities: P1/P2 KL expressions and Bayes recovery quantities;
    \item forecast margins: conversion from structural mismatch to population prediction gaps;
    \item selection stability: concentration error bounds and the $2\tau_{\rm total}$ stability threshold.
\end{enumerate}
The dependency graph is
\[
\text{KL identities}
\rightarrow
\text{forecast margin}
\rightarrow
\text{near-optimal structure}
\rightarrow
\text{structural stability}.
\]

\subsection{Proof of \cref{prop:p1-kl}}

Both candidates are Gaussian with covariance $\sigma^2I$ and means
\[
\mu_d=\alpha S_du,
\qquad
\mu_{d'}=\alpha S_{d'}u.
\]
The equal-covariance Gaussian KL formula gives
\[
D_{\rm KL}(P_d^{(u)}\Vert P_{d'}^{(u)})
=
\frac12
(\mu_d-\mu_{d'})^\top
(\sigma^2I)^{-1}
(\mu_d-\mu_{d'})
=
\frac{\alpha^2}{2\sigma^2}
\norm{S_du-S_{d'}u}_2^2.
\]

\subsection{Proof of \cref{cor:p1-average}}

Stationarity gives
\[
\E[
(U_{t-d}-U_{t-d'})^2
]
=
2\sigma_U^2
[
1-\rho_U(|d-d'|)
].
\]
Summing over $n_{\rm eff}$ positions and using \cref{prop:p1-kl} proves the result.

\subsection{Proof of \cref{prop:binary-bayes}}

Let
\[
\Delta_M^2
=
(\mu_d-\mu_{d'})^\top
\Sigma^{-1}
(\mu_d-\mu_{d'}).
\]
For equal priors and shared covariance, the optimal Gaussian decision rule has error
\[
\Phi(-\Delta_M/2).
\]
Since
\[
D_{\rm KL}(P_d\Vert P_{d'})=\Delta_M^2/2,
\]
the result follows.

\subsection{Proof of \cref{prop:arx-known}}

The conditional trajectory KL is the sum of one-step KL terms. At time $t$, the two conditional means are
\[
a^\star Y_{t-1}
+
b^\star u_{t-d},
\qquad
a^\star Y_{t-1}
+
b^\star u_{t-d'}.
\]
The shared output term cancels, leaving
\[
b^\star
(u_{t-d}-u_{t-d'}).
\]
Summing the equal-variance Gaussian KL terms gives the result.

\subsection{Proof of Proposition \ref{prop:p1-tradeoff}}\label{app:proof prop p1-tradeoff}

By \cref{eq:p1-forecast-margin,eq:p1-kl-main},
\[
D_{\rm KL}(d_0\Vert d_1\mid u)
=
\frac{
n\Delta_n(d_1;d_0,u)
}{
2\sigma^2
}.
\]
Substituting into \cref{eq:p1-bayes-main} gives the Bayes-recovery-error formula.

Under the true delay,
\[
Y
=
\alpha S_{d_0}u+\epsilon,
\qquad
\epsilon\sim N(0,\sigma^2I_n).
\]
Therefore
\begin{align*}
\frac1n
\E
\|Y-\alpha S_{d_1}u\|_2^2
&=
\frac1n
\E
\|
\epsilon+
\alpha(S_{d_0}-S_{d_1})u
\|_2^2\\
&=
\sigma^2
+
\frac{\alpha^2}{n}
\|
(S_{d_0}-S_{d_1})u
\|_2^2\\
&=
\sigma^2+\Delta_n.
\end{align*}
If $\Delta_{n_{\rm eff}}\to0$ and $n_{\rm eff}\Delta_{n_{\rm eff}}\to\infty$, the forecast penalty goes to zero and the Bayes-recovery-error argument goes to $-\infty$.

\subsection{Proof of Proposition \ref{prop:arx-profile-margin}}\label{app:proof prop arx-profile-margin}

Write $n=n_{\rm P2}$. For candidate $d$ and coefficients $\bm\beta$,
\[
Y_t
=
m_t^\star+\epsilon_t,
\qquad
\E^\star[
\epsilon_t\bm x_{t,d}
]=0.
\]
Hence
\begin{align}
\mathcal R_n(d,\bm\beta)
&=
\sigma^2
+
\frac1n
\sum_t
\E^\star
[
(m_t^\star-\bm x_{t,d}^\top\bm\beta)^2
].
\label{eq:p2-risk-decomp-proof}
\end{align}
Define
\[
Q_d
:=
\inf_{\bm\beta}
\sum_t
\E^\star
[
(m_t^\star-\bm x_{t,d}^\top\bm\beta)^2
].
\]
Then
\[
Q_d
=
n\Delta_{\rm prof,n_{\rm P2}}(d).
\]

With fixed candidate variance $\sigma^2$,
\[
D_{\rm KL}(
P^\star
\Vert
P_{d,\bm\beta,\sigma^2}
)
=
\frac1{2\sigma^2}
\sum_t
\E^\star
[
(m_t^\star-\bm x_{t,d}^\top\bm\beta)^2
].
\]
Profiling over $\bm\beta$ gives
\[
J_{\rm prof}^{\rm fixed}
=
\frac{
n\Delta_{\rm prof,n_{\rm P2}}(d)
}{
2\sigma^2
}.
\]

If the candidate uses free shared variance $s^2$, the KL is
\[
\frac12
\left[
n\log\frac{s^2}{\sigma^2}
+
\frac{
n\sigma^2+Q_d
}{s^2}
-n
\right].
\]
The minimizer is
\[
s_{\rm opt}^2
=
\sigma^2+\frac{Q_d}{n}.
\]
Substitution gives
\[
J_{\rm prof}^{\rm free}
=
\frac n2
\log
\left(
1+
\frac{
\Delta_{\rm prof,n_{\rm P2}}(d)
}{\sigma^2}
\right).
\]

\subsection{Derivation of the P2 partial-correlation margin}
\label{app:partialcorr}

We derive the closed-form expression of the profiled forecast margin.
By definition, after re-fitting the coefficients of the alternative delay $d$,
the remaining prediction error is

\[
\Delta_{\rm prof}(d)
=
(b^\star)^2
\inf_{a',b'}
\E^\star
\left[
\left(
U_{t-d^\star}
-a'Y_{t-1}
-b'U_{t-d}
\right)^2
\right].
\]

The linear component explained by $Y_{t-1}$ can be removed from both delayed
inputs. Applying the Frisch--Waugh--Lovell theorem, the above optimization is
equivalent to first removing the contribution of $Y_{t-1}$ and then projecting
the remaining part of $U_{t-d^\star}$ onto the remaining part of $U_{t-d}$.

Therefore, the residual prediction error can be written directly as

\[
\inf_{a',b'}
\E^\star
\left[
\left(
U_{t-d^\star}
-a'Y_{t-1}
-b'U_{t-d}
\right)^2
\right]
=
\operatorname{Var}
\left(
U_{t-d^\star}\mid Y_{t-1}
\right)
\left(
1-
\rho^2_{U_{t-d^\star},U_{t-d}\mid Y_{t-1}}
\right).
\]

Hence,

\[
\boxed{
\Delta_{\rm prof}(d)
=
(b^\star)^2
\operatorname{Var}
\left(
U_{t-d^\star}\mid Y_{t-1}
\right)
\left(
1-
\rho^2_{U_{t-d^\star},U_{t-d}\mid Y_{t-1}}
\right)
}.
\]

This expression shows why re-fitting reduces the forecast margin. The first
factor measures how much of the true delayed input remains unexplained by the
past output. The second factor measures how much of this remaining information
cannot be replaced by the alternative lag.

\subsection{Proof of \cref{prop:forecast-certificate}}

If $\varepsilon<\Delta^{\min}$, every alternative $d$ satisfies
\[
r^\star(d)
>
r^\star(d^\star)+\varepsilon,
\]
so the near-optimal set contains only $d^\star$. If
$\varepsilon\ge\Delta^{\min}$, an alternative attaining the minimum margin belongs to the near-optimal set.

\subsection{Proof of \cref{thm:structural-stability}}\label{app:thm structural-stability}

For any alternative $d$,
\begin{align*}
\widehat r(d)-\widehat r(d^\star)
&\ge
[r^\star(d)-\tau_{\rm total}]
-
[r^\star(d^\star)+\tau_{\rm total}]\\
&\ge
\Delta^{\min}-2\tau_{\rm total}.
\end{align*}
Thus $\Delta^{\min}>2\tau_{\rm total}$ makes every alternative strictly worse.

For sharpness, let $d_1$ attain the minimum margin. Set
\[
\widehat r(d^\star)
=
r^\star(d^\star)+\tau_{\rm total},
\qquad
\widehat r(d_1)
=
r^\star(d_1)-\tau_{\rm total},
\]
and leave the other candidate values unchanged. Then
\[
\widehat r(d_1)-\widehat r(d^\star)
=
\Delta^{\min}-2\tau_{\rm total}.
\]
This is negative below the boundary and zero at equality.

\section{Additional Theoretical Results}
\label{app:additional-theory}

\begin{proposition}[Conditional P1 delay information]
\label{prop:p1-kl}
For any $d,d'\in\cD$,
\[
D_{\rm KL}(P_d^{(u)}\Vert P_{d'}^{(u)})
=
\frac{\alpha^2}{2\sigma^2}
\norm{S_du-S_{d'}u}_2^2.
\]
\end{proposition}

\begin{corollary}[Average P1 information under stationary input]
\label{cor:p1-average}
If $U_t$ is second-order stationary with variance $\sigma_U^2$ and autocorrelation $\rho_U$, then for $r=|d-d'|$,
\[
\E_U
[
D_{\rm KL}(P_d^{(U)}\Vert P_{d'}^{(U)})
]
=
n_{\rm eff}
\frac{\alpha^2\sigma_U^2}{\sigma^2}
[1-\rho_U(r)].
\]
\end{corollary}

\begin{proposition}[Exact binary Bayes recovery error]
\label{prop:binary-bayes}
For two equal-prior Gaussian candidates with the same covariance,
\[
e_{\rm rec}^\star(d,d';u)
=
\Phi\!\left(
-\sqrt{
D_{\rm KL}(P_d^{(u)}\Vert P_{d'}^{(u)})/2
}
\right).
\]
\end{proposition}

\begin{proposition}[Known-parameter P2 delay information]
\label{prop:arx-known}
If the two P2 candidates use the same true coefficients and variance, then
\[
D_{\rm KL}(P_{d,(a^\star,b^\star)}^{(u)}
\Vert
P_{d',(a^\star,b^\star)}^{(u)})
=
\frac{(b^\star)^2}{2(\sigma)^2}
\sum_{t\in\cI_L^{\rm P2}}
(u_{t-d}-u_{t-d'})^2.
\]
\end{proposition}

\begin{proposition}[Near-optimal forecasting and structure]
\label{prop:forecast-certificate}
Let $d^\star$ uniquely minimize the ideal candidate risk and let
\[
\Delta^{\min}
=
\min_{d\ne d^\star}
[r^\star(d)-r^\star(d^\star)].
\]
For
\[
\cA(\varepsilon)
=
\{
d:
r^\star(d)
\le
r^\star(d^\star)+\varepsilon
\},
\]
we have
\[
\cA(\varepsilon)=\{d^\star\}
\quad\text{if}\quad
0\le\varepsilon<\Delta^{\min},
\]
while $\cA(\varepsilon)$ contains an alternative structure when
$\varepsilon\ge\Delta^{\min}$.
\end{proposition}

\begin{proposition}[Partial-correlation form of the profiled margin]
\label{prop:p2-partialcorr-margin}
In the centered stationary Gaussian P2 setting, the reduction of the profiled
forecast margin can be characterized by
\[
\Delta_{\rm prof}(d)
=
(b^\star)^2
\operatorname{Var}
\left(
U_{t-d^\star}\mid Y_{t-1}
\right)
\left(
1-\rho^2_{U_{t-d^\star},U_{t-d}\mid Y_{t-1}}
\right).
\]
The first factor measures the part of the true delayed input that cannot be
explained by the past output. The second factor measures the part of this
remaining signal that cannot be replaced by the alternative lag.
\end{proposition}

This expression explains why re-fitting can reduce the profiled forecast
margin. If the past output or another lag already contains predictive
information about the true delayed input, the alternative structure can
recover part of the missing signal after re-fitting, leaving a smaller
forecast difference.

\section{Bayes Oracle Computation}
\label{app:oracle}

For a finite set of fully specified candidates with prior masses $\pi_j$ and conditional densities $p_j^{(u)}$, the Bayes recovery error is
\[
e_{\rm rec}^\star(u)
=
1-
\int
\max_j
\{
\pi_jp_j^{(u)}(y)
\}
\,dy.
\]
For binary P1, the closed form in \cref{eq:p1-bayes-main} is used. In numerical checks, the MAP rule is also evaluated by Monte Carlo using independently generated candidate labels and Gaussian trajectories. The Monte Carlo standard error of an empirical error rate $\widehat R$ based on $B$ independent decisions is
\[
\sqrt{
\widehat R(1-\widehat R)/B
}.
\]

\section{Experimental Protocols}
\label{app:experimental-protocols}

\subsection{P1 scale study}

The P1 phase study uses
\[
\beta\in\{0.5,1,1.25\},
\qquad
n\in
\{128,256,512,1024,2048,4096,8192\},
\]
with $200$ independent input trajectories per condition and $10{,}000$ MAP decisions per condition. The master seed is fixed. The experiment checks the exact KL identity, analytic/MAP Bayes agreement, and the slopes of $\Delta_n$ and $n\Delta_n$.

\subsection{P2 population margin study}

The P2 population study uses stationary Gaussian AR(1) input and scans
\[
a\in\{0,0.3,0.6,0.85\},
\qquad
\rho\in\{0,0.3,0.6,0.85\},
\qquad
\sigma\in\{1,0.5,0.2\},
\]
with $b=0.5$, true delay $2$, and alternative delay $1$. Stationary covariances are solved analytically by linear state-space equations. The profiled forecast margin is computed independently by three equivalent methods. A finite-sample check uses $200$ trajectories per condition.

\subsection{Controlled P1 learning study}

The controlled P1 learning study keeps accumulated structural information fixed while scanning the per-step margin. A small soft or hard straight-through gate receives only context-based candidate forecast evidence and is trained by forecast MSE. The experiment checks that Bayes/MAP recoverability remains nearly fixed within each information level. Its main result is negative: selection accuracy does not show a stable decrease as the margin becomes smaller.

\subsection{Finite-objective stability study}

The finite-objective study uses $90$ conditions over forecast margin, objective sample size, and noise level, with $5{,}000$ independent replicates per condition. The ideal candidate risks are known exactly, so $\tau_{\rm total}$ is measured directly. A separate deterministic construction checks the sharp boundary at $\Delta=2\tau$.

\subsection{P1 neural candidate study}

The P1 neural candidate study trains separate candidate-restricted neural forecasters and selects between them using independent forecast MSE. Ideal candidate risks are known analytically. This allows the total objective distortion to be measured and also supports a finer decomposition into trained-model and finite-selection contributions.

\subsection{P2 neural stress study}

The P2 stress study uses
\[
a\in\{0,0.6,0.85\},
\quad
\rho\in\{0,0.6,0.85,0.95\},
\quad
\sigma\in\{1,2,4\},
\quad
b\in\{0.5,0.25\},
\]
for $72$ mechanisms. Training sizes are
\[
16,\ 64,\ 256,
\]
selection sizes are
\[
8,\ 32,\ 128,
\]
and eight fixed training seeds are used. This gives $3{,}456$ neural candidate models and matched OLS candidate fits. The final analysis uses $5{,}184$ neural and $5{,}184$ OLS forecast-only structural selections. Total objective distortion is computed directly from the selection losses and the analytic ideal profiled risks; no trained-neural population-risk estimate is required.

\subsection{Why forecast margin alone is insufficient}
\label{sec:margin-analysis}

The stability result does not treat the forecast margin alone as a failure
criterion. Structural selection depends on whether the available separation
remains large relative to the distortion introduced by finite-sample
estimation and model fitting. We therefore examine whether decreasing the
forecast margin by itself leads to more structural errors.

We first use a controlled P1 learner in which the accumulated recovery
information is kept nearly fixed while the per-observation forecast margin
$\Delta$ varies. A small gate receives candidate forecast evidence and is
trained only through forecast MSE. Although the optimization signal becomes
weaker as $\Delta$ decreases, structural accuracy does not exhibit a
corresponding monotone deterioration. We then repeat the analysis with
independently trained P1 neural candidates and compare several quantities with
the observed structural-selection errors.

\begin{table}[htbp]
\tableformat
\caption{
Additional analysis of forecast margin and structural-selection difficulty.
The controlled gate experiment shows that a smaller $\Delta$ weakens the
optimization signal without by itself producing systematic selection failure.
In the neural candidate study, $S_{\rm total}$ has a stronger association with
wrong selection than $\Delta$ alone.
}
\label{tab:margin-analysis}
\begin{tabular}{llc}
\toprule
Study & Quantity & Result \\
\midrule
Controlled gate
& Gradient slope, soft gate
& $0.987$ \\
&
Gradient slope, hard straight-through gate
& $1.01$ \\
\midrule
Neural candidates
& Association: $\log\Delta$
& $-0.159$ \\
&
Association: $\log S_{\rm model}$
& $-0.279$ \\
&
Association: $\log S_{\rm total}$
& $\mathbf{-0.409}$ \\
&
Wrong selections with $S_{\rm total}>1$
& $0$ \\
\bottomrule
\end{tabular}
\end{table}

The controlled experiment separates optimization difficulty from structural
failure. As the forecast margin becomes smaller, the initial gradient
decreases approximately proportionally, but the learner can still select the
correct structure when the remaining distortion is sufficiently small.
Consistently, the neural candidate study shows that $\Delta$ alone has only a
weak association with wrong selection, whereas the association becomes
stronger after model distortion is incorporated into $S_{\rm model}$ and
$S_{\rm total}$.

These results support the interpretation of the stability theorem rather than
introducing a separate selection criterion: a small forecast margin can make
optimization harder, but it is not by itself a sufficient condition for
structural failure. What matters is the separation relative to the total
objective distortion.

\subsection{Data-size effects on objective distortion}

Finally, we examine how the distortion terms change with available data.
Increasing the training size from $16$ to $256$ increases $S_{\rm total}$ in
$1{,}090/1{,}728$ neural comparisons and $1{,}161/1{,}728$ OLS comparisons,
while increasing the selection size from $8$ to $128$ decreases
$\tau_{\rm total}$ in $1{,}373/1{,}728$ neural comparisons and
$1{,}430/1{,}728$ OLS comparisons. These results confirm that the final selection difficulty is controlled by the balance between structural separation and objective distortion.

\subsection{P2: predictive substitution reduces the profiled structural margin}
\label{sec:p2-experiment}

P1 studies structural separation when the model parameters are fixed. In that
setting, the forecast difference is directly determined by the divergence
between the true and alternative structural distributions. P2 considers the
more realistic forecast-only setting, where a predictor is re-fitted after each
candidate structure is selected. The model parameters therefore become
nuisance variables, and the relevant structural separation is no longer the
fixed-parameter divergence but the profiled divergence after parameter
optimization.

We study this effect in stationary P2 systems with Gaussian AR(1) inputs. For
each candidate delay, the predictor parameters are optimized independently.
We compare two forecast margins. The first, $\Delta_{\rm lag}$, corresponds to
using the incorrect lag as a substitute for the true lag while keeping the
remaining structure fixed. The second, $\Delta_{\rm prof}$, allows full
parameter re-optimization and output-history adjustment, corresponding to the
forecast-only structural-selection procedure analyzed in the theory.

A representative condition is
\[
a=0.3,
\qquad
\rho=0.85,
\qquad
\sigma=0.2,
\qquad
b=0.5 .
\]
The resulting margins are

\begin{table}[htbp]
\centering
\caption{P2 predictive substitution and profiled forecast margins.}
\label{tab:p2-margin}
\begin{tabular}{ccc}
\toprule
Quantity
& Interpretation
& Forecast margin
\\
\midrule
$\Delta_{\rm lag}$
& Wrong lag as a predictive substitute
& 0.069375
\\
$\Delta_{\rm prof}$
& Re-fitted predictor under the wrong structure
& 0.046309
\\
\bottomrule
\end{tabular}
\end{table}

The profiled margin is smaller than the lag-only margin because parameter optimization removes part of the structural discrepancy. This reduction is
not a separate source of information loss; rather, it reflects that forecast-only selection observes the profiled structural difference after the model has adapted to each candidate structure.

Across all 48 conditions,
\[
\Delta_{\rm prof}\leq \Delta_{\rm lag}.
\]
The population expressions computed from the Schur complement, the partial-correlation representation, and direct regression agree to numerical precision. Finite-sample estimates also match the population values closely, with a global median relative error of $0.108\%$ and all 48 conditions below $10\%$.

These results confirm the theoretical distinction between fixed-parameter and
profiled structural separation. In forecast-only selection, the effective
identifiability of a temporal structure is governed by the profiled quantity,
because model parameters are optimized jointly with the structural choice.

\section{Semi-synthetic Industrial Dynamics with Controlled Temporal Structure}
\label{app:semi-synthetic-industrial}

This appendix provides an additional experiment using realistic industrial
process dynamics with a controlled temporal structure. The purpose is not to
validate the theoretical results on physical process delays, but to examine
whether the main phenomenon identified in the paper persists when the
background dynamics are obtained from a realistic multivariate process.

We use the normal-operation trajectories from the Tennessee Eastman Process
(TEP) simulator as background dynamics. The original process variables provide
realistic temporal dependence, autocorrelation, and process noise. We select
several input-output pairs and construct semi-synthetic targets by injecting a
known delay:
\[
Y_t^{\rm new}
=
aY_{t-1}^{\rm new}
+
bX_{t-d^\star}
+
\sigma\epsilon_t .
\]
The injected delay $d^\star$ is known by construction and is only used for
evaluation after forecast-only selection. It is never used during training or
candidate selection.

For each task, candidate delays
\[
d\in\{0,\ldots,20\}
\]
are evaluated using the same forecast-only protocol as in the main experiments.
A separate predictor is trained for each candidate delay, and the selected
structure is the one with the lowest independent forecast MSE:
\[
\hat d=\arg\min_d \widehat r(d).
\]
The experiment therefore combines realistic temporal dynamics with a controlled
structural ground truth.

The first study follows the P1 intuition. As the forecast margin between the
true delay and competing delays decreases, the selected structure does not
necessarily become less recoverable. Across different injected delays and noise
levels, smaller per-observation forecast differences can coexist with similar
or improved structural recovery accuracy.

A second study varies noise level, training size, and selection size to examine
when forecast-only structural selection becomes unreliable. The results show
that structural errors increase when the predictive separation becomes harder
to distinguish under finite data, consistent with the theoretical analysis.

\begin{figure}[htbp]
    \centering
    \includegraphics[width=\linewidth]{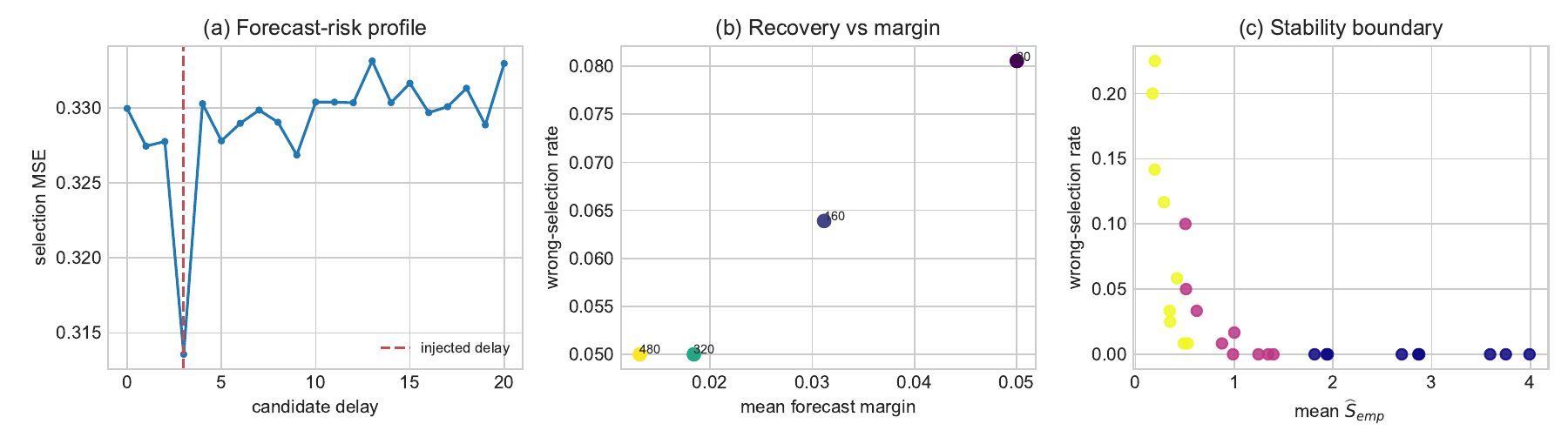}
    \caption{
    Semi-synthetic industrial dynamics experiment. The experiment uses
    Tennessee Eastman Process trajectories as realistic temporal backgrounds
    while injecting controlled temporal structures. The panels show (a) the
    forecast risk profile over candidate delays, (b) representative structural
    selection behavior under repeated evaluation, and (c) the relationship
    between forecast separation and selection consistency.
    }
    \label{fig:semi-synthetic-industrial}
\end{figure}

This experiment should be interpreted as a realistic illustration of the
forecast-only structural selection problem rather than a recovery benchmark for
physical process delays. The structural labels are controlled by construction,
while the temporal background is inherited from an industrial process
simulator.

\section{Additional Empirical Results}

\subsection{P1 neural error sources}

In the P1 neural candidate study, $86.02\%$ of selections are fully correct. Model-level candidate ordering is reversed in $2.82\%$ of cases that remain wrong after finite selection, while $9.95\%$ are finite-selection flips from a correct trained-model ordering. Another $1.20\%$ are accidental corrections of a model-level inversion. The stability guarantee holds in every case with $S_{\rm total}>1$.

\subsection{A milder P2 neural regime}

A separate P2 study with larger training and selection sizes is close to a structural-selection ceiling: $99.74\%$ of selections are correct. The theorem-level condition still holds with zero violations, but the small number of wrong selections gives weak power for comparing different structural scores. This motivates the wider stress grid used in the main P2 experiment.

\subsection{Why predictive substitution is not itself a stability score}

In the P2 stress study, stronger substitution does not have a simple monotone relation with wrong selection. This does not conflict with the population geometry: substitution changes $\Delta_{\rm prof}$, while the final selection also depends on $\tau_{\rm total}$. The experiments therefore use $S_{\rm total}$ as the main stability quantity and treat substitution as one source of margin compression.

\end{document}